# A human-inspired recognition system for pre-modern Japanese historical documents

**Anh Duc Le, Tarin Clanuwat, and Asanobu Kitamoto**

**ABSTRACT** Recognition of historical documents is a challenging problem due to the noised, damaged characters and background. However, in Japanese historical documents, not only contains the mentioned problems, pre-modern Japanese characters were written in cursive and are connected. Therefore, character segmentation based methods do not work well. This leads to the idea of creating a new recognition system. In this paper, we propose a human-inspired document reading system to recognize multiple lines of pre-modern Japanese historical documents. During the reading, people employ eyes movement to determine the start of a text line. Then, they move the eyes from the current character/word to the next character/word. They can also determine the end of a line or skip a figure to move to the next line. The eyes movement integrates with visual processing to operate the reading process in the brain. We employ attention-based encoder-decoder to implement this recognition system. First, the recognition system detects where to start a text line. Second, the system scans and recognize character by character until the text line is completed. Then, the system continues to detect the start of the next text line. This process is repeated until reading the whole document. As results, the system is successful to recognize multiple lines, connected and cursive characters without performing character/line segmentation. Besides, we also employ a coverage model which stores the history of eyes movement to predict the next movement more precisely. We tested our human-inspired recognition system on the pre-modern Japanese historical document provide by the PRMU Kuzushijiji competition. The results of the experiments demonstrate the superiority and effectiveness of our proposed system by achieving Sequence Error Rate of 9.87% and 53.81% on level 2 and level 3 of the dataset, respectively. These results outperform to any other systems participated in the PRMU Kuzushiji competition.

**INDEX TERMS** A human reading-inspired recognition system, recognition of pre-modern Japanese historical document, attention-based encoder-decoder, Kuzushiji.

## I. INTRODUCTION

Through the development of human civilizations, writing systems have been changed over time in every languages. However, in some countries, the changes are so drastic so that younger generations can't read their older languages anymore. For example, Vietnam switched writing system Chinese-based writing system (Chữ nôm) to a Latin-based one. China changed traditional Chinese characters to simplified Chinese characters in 1964. These countries have been suffered from the problem of translating past knowledge written in historical resources to current languages. Since only trained experts can read historical documents, the transcription process is very slow and time-consuming.

Japan had been using Kuzushiji or cursive writing style shortly after Chinese characters got in to the country in 8th century. The Japanese has been using 3 types of characters which are Kanji (Chinese character in Japanese language), Hiragana and Katakana. Hiragana and Katakana; derived from different ways of simplifying Kanji, don't contain independent semantic meaning, but instead carry phonetic information (like letters in the English alphabet).

Because Kuzushiji character were written in cursive, there are many ways in writing one character. One characteristic of classical Japanese which is very different from modern one is that Classical Japanese contains Hentaigana (variant Kana character). Hentaigana is Hiragana characters that have more than one form of writing, as they were derived from different root Kanji. Figure 1 shows shape variations of Hiragana character "ka" which derived from two different root Kanji.

We have a lot of digitized documents collected in libraries and museums throughout the country. However, the images were taken from a few hundred years old books. Therefore, the backgrounds are noised from insect bites or stains simply because they are very old. Some pages are not uniform because of patterns from paper dyeing as shown in Figure 2. Since Kuzushiji characters have such unique characteristics, hence, recognition task using machine is also extremely difficult.

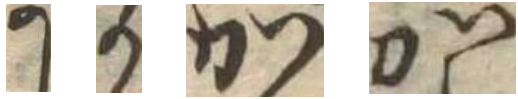

**Figure 1.** Examples of different shapes for the character "ka".

As an aid for learning Kuzushiji, Hashimoto et al. developed a mobile learning application for reading Kuzushiji characters and documents [1]. The application has three modules: character reading, text reading and community modules. The character module is flashcard-like features for users to learn character shapes of 102 hentaigana and 176 kanji characters. The reading module helps users to learn how to read Kuzushiji characters from real classical texts. The classical texts are accompanied by transcriptions. Users can look up the transcriptions when they cannot read the text. The community module let users communicate with each other to exchange learning materials and learning experiences. The application is available on Android and iOS platform and has already been downloaded more than 50,000 times.

Kuzushiji workshops are also regularly held at various places in Europe since 2011 by the National Institute of Japanese Literature and the European Association of Japanese Resource Specialists [2]. The aim of the workshop is to help librarians, scholars and graduate students to obtain practical knowledge and first-hand experience on reading Kuzushiji.

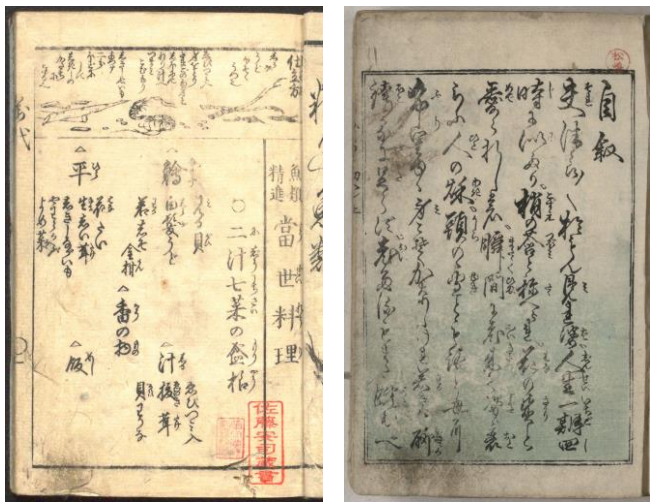

**Figure 2.** Two examples of stained Kuzushiji documents.

For recognizing Kuzushiji characters, Horiuchi et al. [3] employed modular neural networks which consist of a rough-classifier and a set of fine-classifiers to recognize Kuzushiji. They employed the Multi-Templates Matching for the rough-classifier and Multilayered Perceptrons for the fine-classifier — first, the rough-classifier select several candidates of character categories for an input pattern. Then, the fine-classifier finds the best result from the candidates.

In 2017, The Center for Open Data in the Humanities (CODH) in Japan provided Kuzushiji dataset to organize a competition, the 21th Pattern Recognition and Media Understanding (PRMU) Algorithm Contest for Kuzushiji recognition [4]. The competition has three levels: level 1: recognizing isolated Kuzushiji characters, level 2: recognizing single line of three Kuzushiji characters and level 3: recognizing multiple lines of Kuzushiji characters. The organizers provided documents, ground truth of bounding boxes ,and target character codes for three tasks. Nguyen et al. won the competition. They developed three recognition systems based on convolutional neural network (CNN) and Bidirectional Long Short-Term Memory (BLSTM) for three tasks [5]. For recognizing isolated Kuzushiji characters, they employed CNN and 2DBLSTM based methods. For recognizing single line and multiple lines of Kuzushiji characters, they combined architecture of CNN and BLSTM.

Traditional character recognition systems are divided into two main steps: text line detection and text line recognition. In text line detection, documents are decomposed into text lines. Then, each text line is recognized by a text line recognition. The traditional recognition system has good performance on printed and handwritten documents. However, they are still insufficient for historical documents like Kuzushiji documents. They face the difficulty of dealing with cursive and connected characters. Figure 2 shows a page of Kuzushiji document whose characters are connected by writing styles.

In traditional character recognition approach, text lines and characters must be segmented before the system recognizes them. Therefore, errors in the text line detection or character segmentation will be propagated to the recognition step.

To overcome the above problems, we propose a human-inspired historical document reading system. People read documents by scanning characters from line to line and from word to word. They are able to skip figures or spaces and move to the next line. In order to implement this method, we employ the attention-based encoder-decoder model for modeling human reading behavior into the recognition system. The system is able to detect the start character of a line, move to the next character, detect the end of a line to move to the next line, and recognize character at the current attention. The system is trained end-to-end which yield better performance than the previous systems trained separately.

## II. RELATED WORKS

For historical text line detection, C. Clausner et al. proposed a hybrid text line segmentation method based on a combination of rule-based grouping of connected components (bottom-up) and projection profile analysis (top-down). They showed the effectiveness of the proposed method on a diverse set of historical documents in the IMPACT project [6]. Y. Xu et al. proposed a multi-task layout analysis method that uses a single Fully Convolutional Network to perform page segmentation, text line segmentation, and baseline detection on medieval manuscripts, simultaneously [7].

For text line recognition, Segmentation based methods, which attempt to split text line into characters at their true boundaries and label the split characters by using Hidden Markov Model [8] or CNN [9, 10]. Segmentation free method show their advantages in the problems of the sequence to sequence such as handwritten recognition [11], speech recognition [12]. For historical documents, A. Fischer et al. employed two state-of-the-art recognizers for modern scripts (HMM and LSTM) to recognize medieval documents [13].

For recognizing a page of documents, most of the works train individual text line detection, segmentation, recognition and then combine them into a recognition system. D. Wang et al. proposed a real-time recognition to recognize Chinese handwritten pages [14]. Dynamic text line segmentation and character over-segmentation are performed separately to determine a sequence of primitive segments in the page. Text line recognition is integrated with linguistic context and geometric context to provide target text. B. Moysset proposed a recognition system which contains Fully Convolutional Network based text localization network and Multidimensional Long Short-Term Memory based text recognition [15]. C. Wiginton proposed a page handwritten recognition composing Region Proposal Network to find the start position of text lines, line follower network to normalize text lines and CNN-LSTM network to do text recognition [16]. Individual networks are pre-trained separately and then jointly trained together.

Recently, an attention-based encoder-decoder model has been successful in many domains such as machine translation [17], image caption generation [18], and handwriting recognition [19, 20]. It outperforms traditional methods by the power of attention mechanisms in capturing contextual dependencies. The system is easy to train because it requires only images and corresponding transcriptions without character annotation.

## III. A HUMAN-INSPIRED RECOGNITION SYSTEM

### A. HUMAN READING BEHAVIOR

Human reading is involved in eyes movement and visual processing of written text. Eyes movement includes rapid movements (saccades) intermingled with shortstops (fixations). During the reading, people can determine the start character of a paragraph/line. Then, they move the eyes from the current character/word to the next character/word. They can also determine the end of a line or skip a figure to move to the next line. Figure 3 visualizes eye-movement of a person when he/she reads. The fixations are red circles while saccades are dotted lines.

### B. PROBLEM DEFINITION

Recognition of historical documents can be formulated as a sequence labeling problem. Suppose we have an image I of historical documents and the corresponding ground truth $s = \{s_1, s_2, ..., s_N\}$ consisting of $N$ character. Our task is to predict a sequence of characters $y = (y_1, y_2..., y_N)$. The possible characters $y_i \in Y$ at time-step i are a set of vocabulary. This set of vocabulary $Y$ contains special tokens that denote a start token ($<S>$), an end of sentence token ($<E>$). Given a training set $D = \{(I, s)\}$ which contains pairs$(I, s)$, we try to maximize w.r.t. parameters $\theta$ of the recognition system $p(y_1, y_2,..., y_N / I, \theta)$. In this research, the proposed recognition system is based on the attention-based encoder-decoder model.

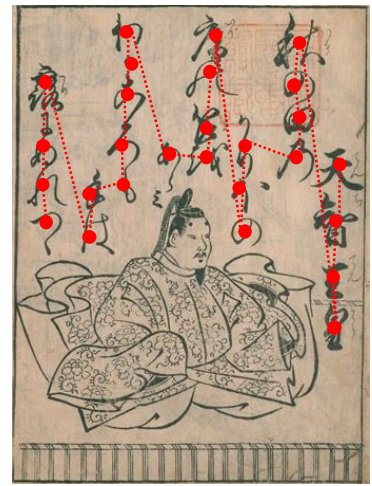

**Figure 3.** An example of eyes movement during reading.

### C. ATTENTION-BASED ENCODER-DECODER

Our recognition system is based on the attention-based encoder-decoder approach. The architecture of our recognition system is shown in Figure 4. It contains two modules: a convolution neural network for feature extraction and an LSTM Decoder with an attention model for generating the target characters. These modules are described in the following sections.

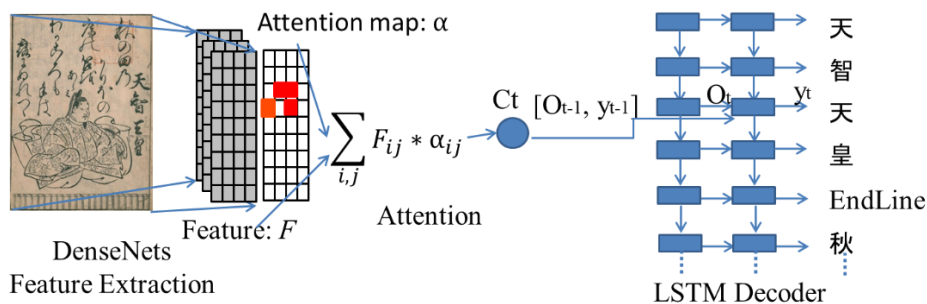

Figure 4. The architecture of the attention-based encoder-decoder.

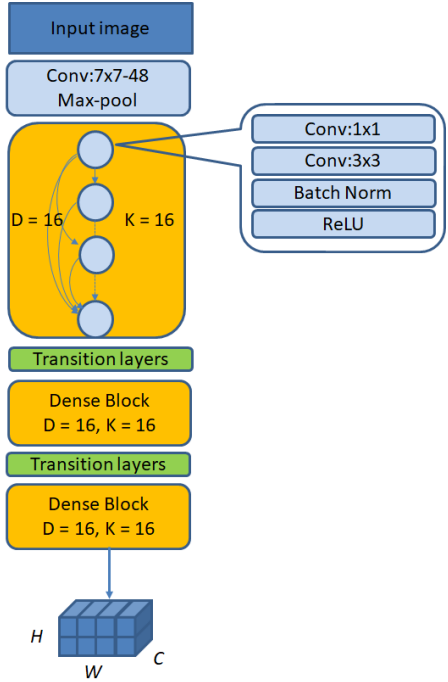

Figure 5. The architecture of DenseNets for feature extraction.

**Encoder:** The related works on image classification [9] and handwritten math recognition [13] have been verified DenseNets outperforms the VGG and ResNets by proposing direct connections from any preceding layers to succeeding layers. The $i^{th}$ layer receives the feature-maps of all preceding layers, $x0, \ldots, xi-1$, as input:

$$x_i = H_i([x_0, x_1, \ldots, x_{i-1}]) \quad (1)$$

Where $H_i$ refers to the convolutional function of $i^{th}$ layer and $[x_0, x_1, \ldots, x_{i-1}]$ refers to the concatenation of the output of all preceding layers. Densely connections help the network reuse and learn features cross layers. However, densely connections require more memory because of the growth of connections through depth. To limit connections and keep the same input size, DenseNets is divided into densely connected blocks as Figure 5. In each dense block, we add a blottleneck layers (1x1 convolution layer) before 3x3 convolution layer to reduce the computational complexity. The dense blocks are connected by transition layers which contain convolutional and average pooling layers. For compression, the transition layer reduces a haft of feature maps. The detailed implementation is described as follows. We employ a convolutional layer with 48 feature maps and a max pooling layer to process input image. Then, we employ three dense blocks of growth rate (output feature map of each convolutional layer) $K$ and the depth (number of convolutional layers in each dense block) $D$ to extract features. The size of outputthe features is $HxWxC$.

**Attention-based Decoder:** The input of the decoder can be shown as the following:

$$F = \{F_{u,v}, 0 \le u \le H, 0 \le v \le W, F_{u,v} \in R^C\} \quad (2)$$

The LSTM decoder generates one character at a time. At each time step $t$, the decoder predicts symbol $y_t$ based on the embedding vector of the previous decoded symbol $E_{yt-1}$, the current hidden state of the decoder $h_t$, and the current context vector $c_t$ as the following equation:

$$p(y_t | y_1, \ldots, y_{t-1}, F) = softmax(W(E_{y_{t-1}} + W_h * h_t + W_c * c_t)) \quad (3)$$

The hidden state is calculated by an LSTM. It is based on the previous hidden state of the LSTM, the context vector, and the embedding vector of the previous decoded symbol $E_{yt-1}$ as the following equation:

$$h_t = LSTM(h_{t-1}, c_t, E_{y_{t-1}}) \quad (4)$$

The contex vector $c_t$ is computed by the attention mechanism. It is a result of a weighted sum of the input features $F$ and the attention probability $\alpha$.

$$c_t = \sum_{u,v} \alpha_{t(u,v)} * F_{u,v} \quad (5)$$

$$\alpha_{t(u,v)} = \frac{exp(e_{t(u,v)})}{\sum_{i,j} exp(e_{t(i,j)})} \quad (6)$$

Where $e_{t(u,v)}$ denotes the energy of $F_{u,v}$ at time step t. The energy is calculated from the input feature $F_{u,v}$, the previous hidden state of the decoder $h_{t-1}$, and coverage vector $Cov_t(u,v)$. The coverage vector helps the recognition system to remember previous attention positions. This looks like the way human beings remember scanned positions in a document that they read through already. The coverage vector is initialized as a zero vector and we compute it based on the summation of all past attention probabilities $\alpha$.

$$e_{t(u,v)} = v_{att}^T * tanh(W_h * h_{t-1} + W_F * F_{u,v} + W_{cov} * Cov_{t(u,v)}) \quad (7)$$

$$Cov_{t(u,v)} = \sum_{l=1}^{t-1} \alpha_{l(u,v)} \quad (8)$$

We initialize the hidden of the decoder by zero vector. The character generation process is repeated until the decoder produces <end> character.

**Training:** We employ cross-entropy as the objective function to maximize the probability $p$ of predicted characters as follows:

$$f = \sum_{i=1}^{|D|} \sum_{t=1}^{|L_i|} -\log p(g_{i,t}|y_1, \ldots, y_{t-1}, F_i) \quad (9)$$

where $D$ is the training set, $S_i$ is a training sample, $s_{i,t}$ is a ground truth of the $t^{th}$ character of training sample $S_i$, and $F_i$ is the output of DenseNets of the sample $S_i$.

We use the AdaDelta algorithm with gradient clipping to learn the parameters. The AdaDelta hyperparameters are set as $\rho = 0.95, \varepsilon = 10^{-8}$ and batch size is set to 8. The training process is stopped when the expression rate on the validation set does not improve after 15 epochs. The parameters for DenseNets are set as described in section A. we set the size of LSTM decoder to 256.

## IV. EVALUATION

### A. DATASET

As mentioned earlier, the competition has three tasks for three different difficulty levels. Level 1: recognition of single segmented characters, level 2: recognition of three vertically Kana characters and level 3: recognition of unrestricted sets of three or more than three characters possibly in multiple lines [11]. In this research, we only focused on level 2 and 3 which contain multiple characters and lines. However, our proposed model is able to apply for recognizing a whole page of Kuzushiji. Figure 6 shows a sample page of Kuzushiji documents and examples of three levels. Three tasks require to recognize 46 Hiragana characters, but Kanji and Katakana are not included.

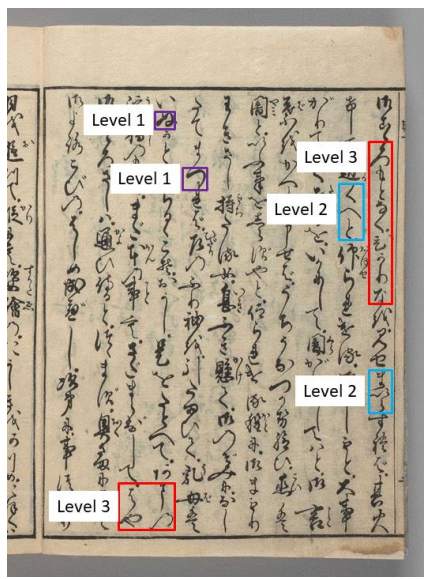

FIGURE 6. A sample page of Kuzushiji documents with the bounding boxes of three levels.

TABLE I
STATISTICS OF THE TRAINING, VALIDATION, AND TESTING SETS FOR LEVEL 2 AND 3

|  | Level 2 | Level 3 |
| --- | --- | --- |
| Training set | 56,097 | 10,118 |
| Validation set | 6,233 | 1,125 |
| Testing set | 16,835 | 1,340 |

The dataset is constructed from 2,222 scanned pages of 15 historical books provided by CODH. Since the testing set in the PRMU competition has not released yet, we separated the dataset into training, validation, and testing set as the winning team of the competition [5]. All the samples in the 15[th] book, named as *Hiyokurenri Hananoshimadai* (比翼連理花洒志満台), are selected for the testing set of level 2, and 3. The remaining samples are divided randomly into the training and validation sets with the ratio of 9:1. Table I shows the statistics of the training, validation, and testing sets for level 2 and 3.

### B. EVALUATION METRICS

In order to measure the performance of our proposed system on level 2 and 3, we use the Character Error Rate (CER) and Sequence Error Rate (SER) metrics which is generally employed for evaluating handwriting recognition systems. The detail of the metrics are shown in the following equations:

$$CER(T, h(T)) = \frac{1}{Z} \sum_{(I,s) \in T} ED(s, h(I)) \quad (10)$$

$$SER(T, h(T)) = \frac{1}{|T|} \sum_{(I,s) \in T} \begin{cases} 0 \ if \ s = h(I) \\ 1 \ otherwise \end{cases} \quad (11)$$

Where $T$ is a testing set which contains input-target pairs ($I$, $s$), $h(I)$ is the output of a recognition system, $Z$ is the total number of target character in $T$ and $ED(s, h(I))$ is the edit distance function which computes the Levenshtein distance between two strings $s$ and $h(I)$.

### B. EXPERIMENTS

The first experiment was for selecting the best hyperparameters (the growth rate $K$ and the depth $D$) of the DenseNets feature extraction. Since the structure of feature extraction affects the performance of the end-to-end system [8-11], we tried four settings of the hyperparameters to train the recognition system on the training and validation sets of level 3. We employ three densely blocks as Figure 4 and change the number of growth rate and depth. The result is shown in Table II. The best hyperparameter was selected based on the CER on the validation set of level 3, which was $K = 16$, and $D = 16$. We employ the best hyperparameters for the remaining experiments.

The second experiment evaluated the performance of the proposed recognition system on the validation set of level 2 and 3 and compared with other methods. We employed the training sets as described in Table III. The result is shown in Table III. The proposed recognition system achieved 5.40% of CER and 0.87% of SER by training on level 2 data set and 13.07 of CER and 53.81 of SER by training on level 3. We compare our proposed recognition system with other systems participated in the PRMU Kuzushiji competition. Deep Convolutional Recurrent Network (DCRN) is a combined architecture of CNN, BLSTM, and CTC. VGG_2DBLSTM is a combined architecture of VGG feature extraction and 2-dimensional BLSTM. Faster R-CNN is an object detection to detect and recognize characters in an input image. We employ the results from Nguyen et al.'s work for the comparison. Our proposed human-inspired recognition system outperforms the best system provided by Nguyen et al. (DCRN system is the winner for the Kuzushiji competition). For level 2, our proposed system outperforms the DCRN system 7.48% of CER (5.40% vs. 12.88%) and 21.73% of SER (9.87% vs. 31.60%). For level 3, our proposed system outperforms the DCRN system 13.63% of CER (13.07% vs. 26.70%) and 28.76% of SER (53.81% vs. 82.57).

TABLE II
COMPARISON OF THE PERFORMANCE OF DIFFERENT HYPERPARAMETERS ON THE VALIDATION SETS OF LEVEL 3

|  | DenseNets 1 K = 16, D = 8 | DenseNets 2 K = 24, D = 8 | DenseNets 3 K = 16, D = 16 | DenseNets 4 K = 24, D = 16 |
|---|---|---|---|---|
| CER | 6.78 | 5.23 | 3.74 | 4.06 |
| SER | 23.02 | 19.68 | 15.11 | 17.07 |

TABLE III
COMPARISION OF THE PROPOSED RECOGNITION SYSTEM AND OTHER SYSTEMS ON THE VALIDATION SET OF LEVEL 2 AND 3

|  | Level 2 | | Level 3 | |
|---|---|---|---|---|
|  | CER | SER | CER | SER |
| Proposed system | 5.40 | 9.87 | 13.07 | 53.81 |
| DCRN | 12.88 | 31.60 | 26.70 | 82.57 |
| VGG_2DBLSTM | N/A | N/A | 44.18 | 97.16 |
| Faster R-CNN | N/A | N/A | 60.30 | 99.85 |

## C. ANALYSIS

Figure 7 shows the recognition process of the human-inspired recognition system. The Kuzushiji document has 2 lines. The reading order of line is from right to left. The recognition results are shown at the top of the documents. The attention map is visualized in red color. The blue boxes are bounding boxes of Kuzushiji characters. It is created manually by a Kuzushiji's expert. The recognition system recognized Kuzushiji characters correctly, and the attention model focused correctly on the positions of characters. The recognition process is stopped when it reaches <End> symbol. The dataset does not provide the bounding box for each character. We cannot evaluate the precise of attention map. However, we observed that the attention model provide very precise attention positions when the system recognizes by visualizing the recognition process for 10 Kuzushiji documents.

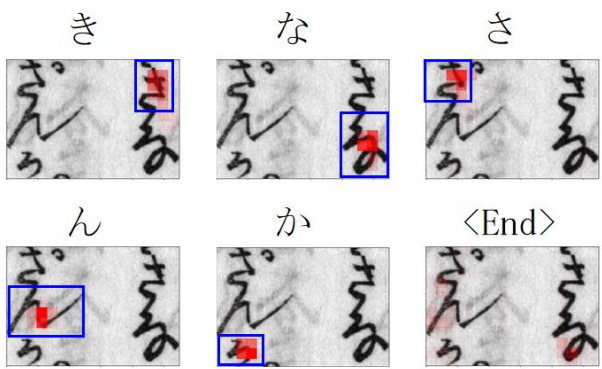

**FIGURE 7.** Visualizing the recognition process of the human-inspired recognition system for a Kuzushiji document. Note that Kuzusiji is read vertically from right to left.

Figure 8 shows two examples of correct and incorrect recognition results. The blue boxes are provided by Kuzushiji's expert. The black characters are correct recognition results while red characters are incorrect recognition results. The results show our proposed system is effective in dealing with cursive and connected characters.

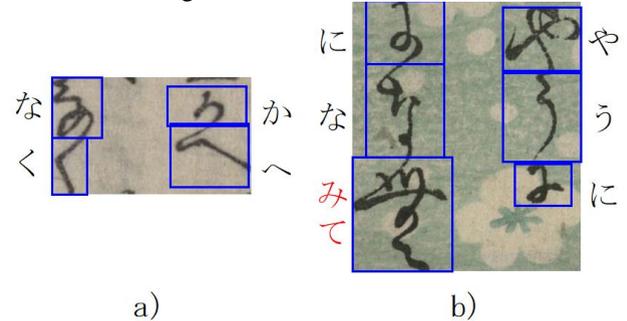

**FIGURE 8.** Examples of corrected and incorrected recognition results.

## V. CONCLUSION

In this paper, we introduce a human-inspired recognition system for recognizing Kuzushiji documents. Our system simulates the human reading behavior which is able to determine the start character of a text line, scan the next character and determine the end of a text line to move to the next text line. The system performs much better than the winner of the 2017 PRMU Algorithm Contest in Kuzushiji recognition. The system achieved 15.2% of 60.3% of SER on the testing set of level 2 and 3, respectively. We also visualized the attention map and observed the attention model provide very precise attention location.

In the future, we plan to expand the human-inspired recognition system to recognize the whole page of Kuzushiji documents. We also plan to integrate the current system with unsupervised learning to reuse unlabeled document and improve the performance of the recognition system.

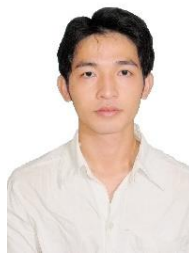
**Anh Duc Le** was born on 8 April 1988 in Vietnam. He received his B.Eng. degree in Computer Science from honor program, Ho Chi Minh City University of Technology. He received his Master and Ph.D. degree from Tokyo University of Agriculture and Technology in 2014 and 2017, respectively. He is currently working on handwritten recognition, pattern recognition, deep learning, and their applications.

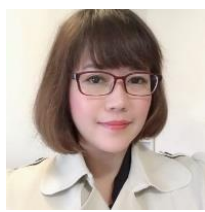
**Tarin Clanuwat** received her master and Ph.D. in Japanese literature from Waseda University, Tokyo. She specialized in the Tale of Genji commentary books from 12th to 15th century. She is currently working on Kuzushiji recognition system with deep learning and released dataset for machine learning benchmarks such as Kuzushiji-MNIST.

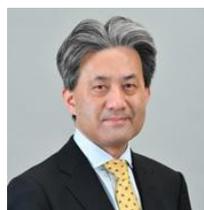
**Asanobu Kitamoto** earned his Ph.D. in electronic engineering from the University of Tokyo in 1997. He is now Director of Center for Open Data in the Humanities (CODH), Joint Support-Center for Data Science Research, Research Organization of Information and Systems (ROIS), Associate Professor of National Institute of Informatics, and SOKENDAI (The Graduate University for Advanced Studies). His main technical interest is image processing, but he also extends the approach of data-driven science into a wide range of disciplines such as humanities, earth science and environment, and disaster reduction. He received Japan Media Arts Festival Jury Recommended Works, IPSJ Yamashita Award, and others. He is also interested in trans-disciplinary collaboration for the promotion of open science.